\documentclass[letterpaper, 10 pt, conference]{ieeeconf}  

\IEEEoverridecommandlockouts                              

\usepackage{graphics} 
\usepackage{epsfig} 
\usepackage{mathptmx} 
\usepackage{times} 
\usepackage{amsmath} 
\usepackage{amssymb}  

\usepackage{booktabs}
\usepackage{dsfont}

\usepackage{bbding}
\usepackage{pifont}
\usepackage{wasysym}
\usepackage{color, colortbl}
\usepackage{eucal}

\usepackage{algorithm}
\usepackage{algpseudocode}
\usepackage{stackengine}
\usepackage{multirow}
\usepackage{tabularx}
\usepackage{subcaption}
\usepackage{esvect}
\usepackage{makecell}
\usepackage{overpic}
\usepackage{wrapfig}
\usepackage{tikz, xcolor}
\usetikzlibrary{calc,spy}
\usepackage{hyperref}
\usepackage{cleveref}
\crefname{figure}{Fig.}{Figures}

\title{\LARGE \bf
Direct Preference Optimization-Enhanced Multi-Guided Diffusion Model for Traffic Scenario Generation
}

\author{Seungjun Yu$^{1\dagger}$, Kisung Kim$^2$, Daejung Kim$^{2}$, Haewook Han$^{1}$, Jinhan Lee$^{2*}$%
\thanks{*This research was supported by a grant from Innovation Project for Autonomous Driving Technology Development funded by Ministry of Land, Infrastructure and Transport of Korean government~(No.~2610000148).} 
\thanks{$^{1}$Seungjun Yu, Haewook Han are with School of Electrical Engineering, Pohang University of Science and Technology, Pohang, Republic of Korea {\tt\small \{seungjunyu, hhan\}@postech.ac.kr}}%
\thanks{$^{2}$Kisung Kim, Daejung Kim, Jinhan Lee are with Autonomous Driving Group, NAVER LABS {\tt\small \{ks.kim, daejung.kim, jinhan.lee\}@naverlabs.com}
}%
\thanks{$^*$Corresponding author.}%
\thanks{$\dagger$ Work done during an internship at NAVER LABS.}
}

\begin{document}

\maketitle
\thispagestyle{empty}
\pagestyle{empty}

\begin{abstract}
Diffusion-based models are recognized for their effectiveness in using real-world driving data to generate realistic and diverse traffic scenarios. These models employ guided sampling to incorporate specific traffic preferences and enhance scenario realism. However, guiding the sampling process to conform to traffic rules and preferences can result in deviations from real-world traffic priors and potentially leading to unrealistic behaviors. To address this challenge, we introduce a multi-guided diffusion model that utilizes a novel training strategy to closely adhere to traffic priors, even when employing various combinations of guides. This model adopts a multi-task learning framework, enabling a single diffusion model to process various guide inputs. For increased guided sampling precision, our model is fine-tuned using the Direct Preference Optimization (DPO) algorithm. This algorithm optimizes preferences based on guide scores, effectively navigating the complexities and challenges associated with the expensive and often non-differentiable gradient calculations during the guided sampling fine-tuning process. Evaluated using the nuScenes dataset our model provides a strong baseline for balancing realism, diversity and controllability in the traffic scenario generation.
Supplementary videos are available on the \url{https://sjyu001.github.io/MuDi-Pro/}

\end{abstract}

\section{INTRODUCTION} \label{sec:intro}

The generation of realistic, diverse, and controllable traffic scenarios is crucial for testing and ensuring the safety of autonomous vehicles~\cite{review}. However, acquiring such scenarios from real-world driving is both costly and dangerous~\cite{intro1}, leading to a demand for the automatic generation of synthetic scenarios.  Recent research has responded to this need by combining a learning diffusion-based traffic model, grounded in real-world driving data, with a heuristic-driven approach known as guided sampling. This combination has yielded an effective balance between realism and controllability in scenario generation. Nonetheless, a trade-off between these two aspects remains, as samples generated through guided sampling often diverge from the data distribution and exhibit unrealistic behaviors \cite{ctg+llm, trafficRLHF, realgen}. 
%
%
The challenge in developing a traffic model lies in striking a balance between realism, diversity, and controllability in traffic simulations. CTG++~\cite{ctg+llm} seeks to address this issue by enhancing the robustness of guidance functions through a combination of large language models and data-driven models, utilizing a diffusion transformer. RealGen~\cite{realgen} tackles the problem by integrating retrieval-augmented generation with an autoencoder model. However, these methods do not fundamentally resolve the inherent issues in guided sampling. SceneControl~\cite{scenecontrol} aims to create realistic scenarios by implementing a robust diffusion framework effectively. TrafficRLHF~\cite{trafficRLHF} focuses on improving realism (i.e., reducing unrealistic collisions between vehicles and off-road driving) by fine-tuning a pre-trained diffusion model through reinforcement learning with human feedback. This approach requires collecting human evaluations of traffic scenarios to train a reward model.\\
\begin{figure}[t]
  \centering
  \includegraphics[width=\linewidth]{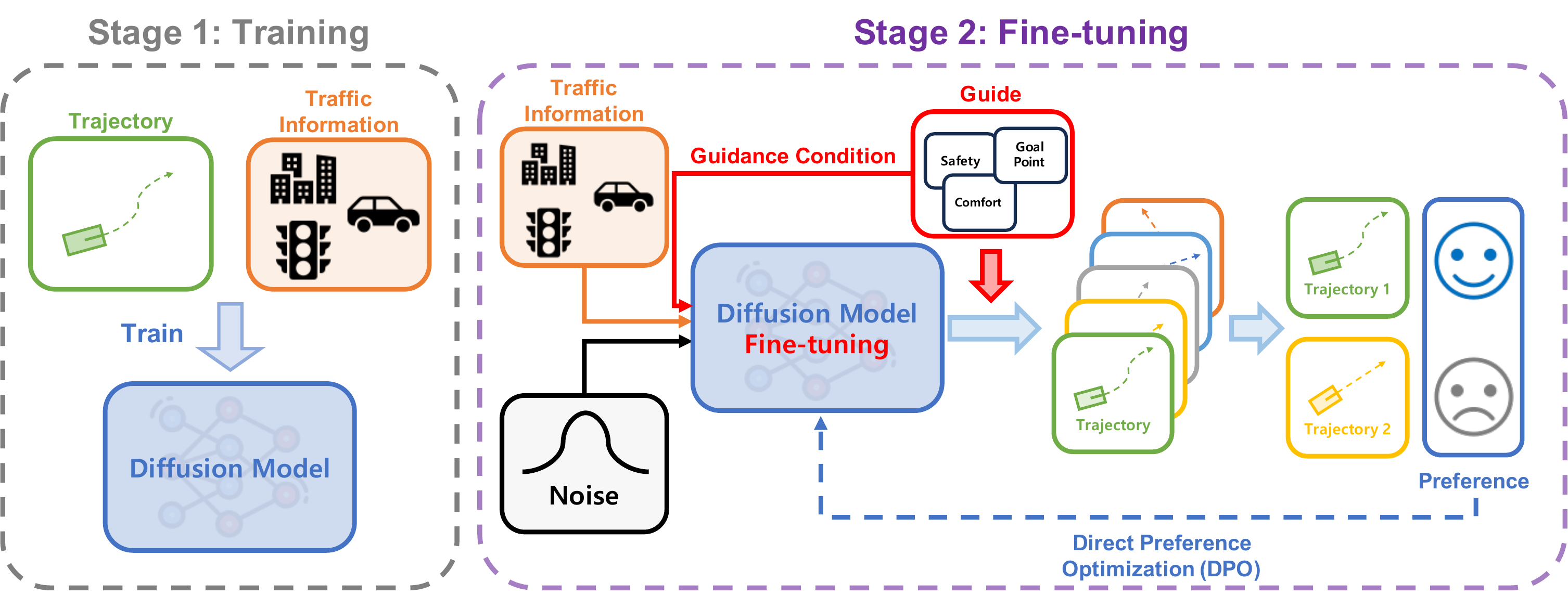}

  \caption{\textbf{Overview of MuDi-Pro Training Process.} MuDi-Pro employs a two-step training process.}
  \label{fig1}
  \vspace{-22pt}
\end{figure}
To generate samples that adeptly balance realism, diversity, and controllability, we introduce a novel methodology: the Multi-Guided Diffusion Model with Direct Preference Optimization (MuDi-Pro). This approach employs a new training strategy inspired by Direct Preference Optimization (DPO)~\cite{dpo} and adopts multi-task learning principles.
MuDi-Pro incorporates three core technical ideas: (1) Utilizing diffusion transformers and classifier-free sampling to learn a traffic prior at the scene level, thereby training a robust data-driven backbone model. The diffusion transformer is trained with real-world driving data (\cref{fig1}, Left). (2) Introducing a guidance conditional layer, analogous to the task conditional layer in multi-task learning \cite{taskconditional, multitaskdiffusionfinetuning}, specifies the appropriate guidance for guided sampling enabling a single model to adeptly manage various combinations of guides. (3) Fine-tuning the pre-trained diffusion model by optimizing it based on preferences over inference samples using guidance scores, rather than relying on human preference feedback (\cref{fig1}, Right).
\begin{figure*}[t]
    \vspace{2mm}
    \centering
    \begin{subfigure}{.69\textwidth}
        \centering
        \includegraphics[height=5cm]{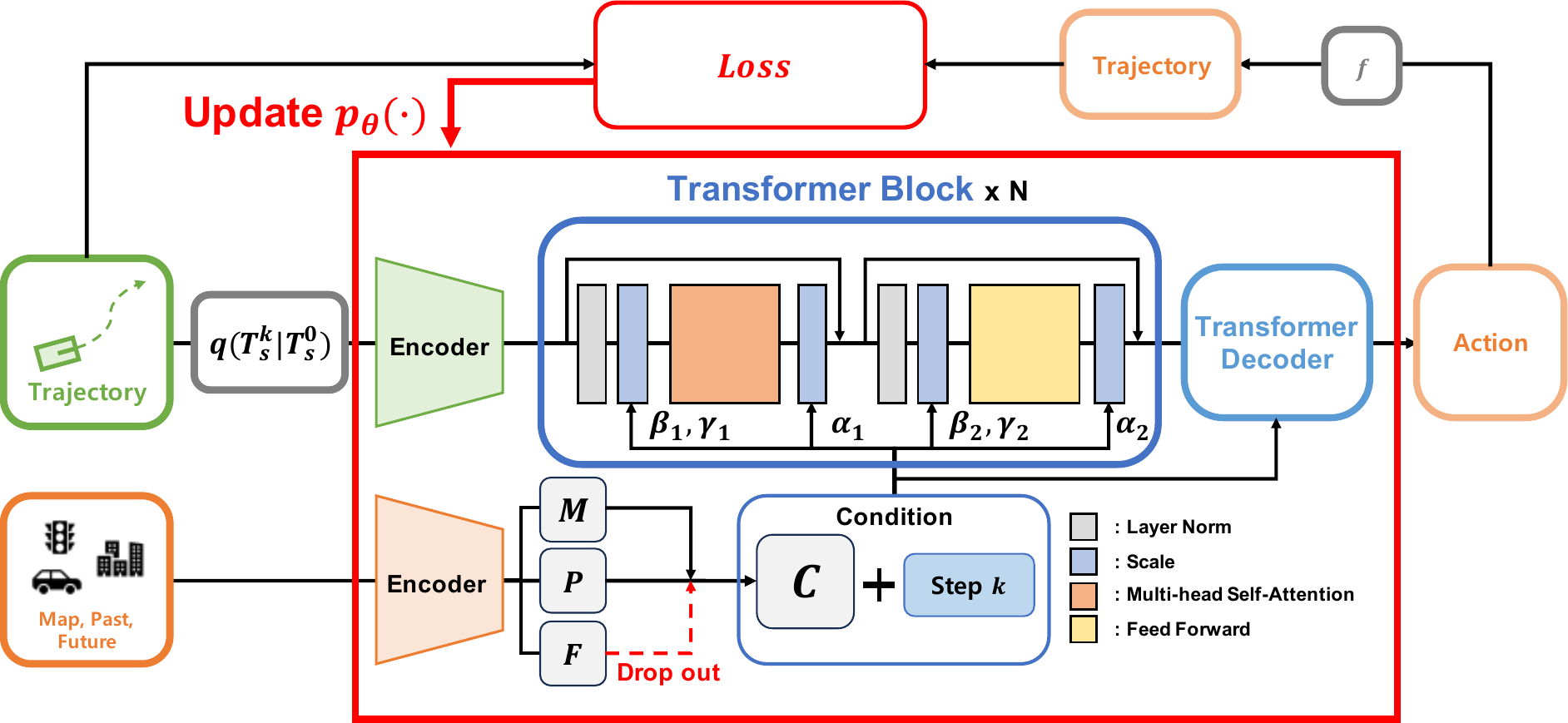}
        \caption{Diffusion Training}
        \label{figure2-a}
    \end{subfigure}
    \hfill
    \begin{subfigure}{.25\textwidth}
        \centering
        \includegraphics[height=5cm]{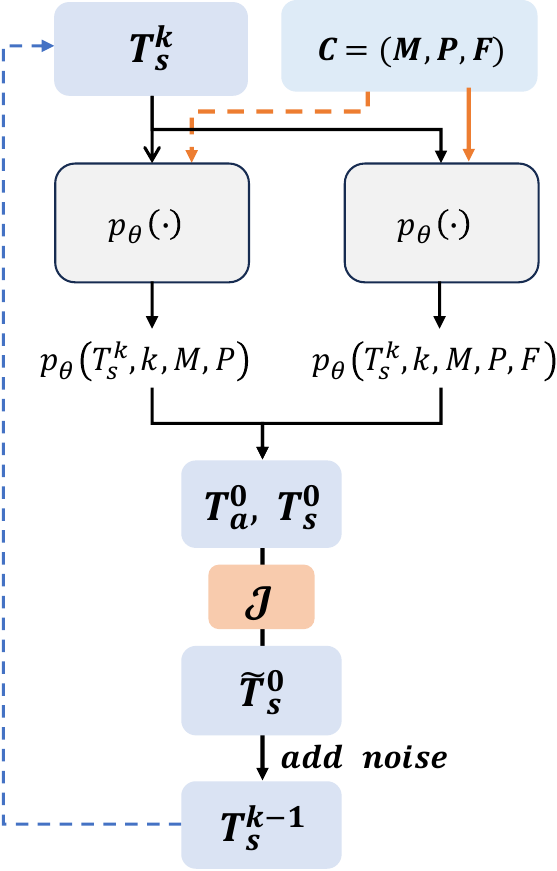}
        \caption{Inference}
        \label{figure2-b}
    \end{subfigure}
    \caption{{\bf Model architecture of training and inference.} (a) The model is trained to predict clean trajectories from noisy ones. The input trajectory and traffic information are first encoded, then processed through Transformer blocks. (b) In the inference phase, classifier-free sampling controls the amount of future feature information provided to the sample, while guided sampling directs the sample generation towards the desired outcome during the denoising process.}
    \label{fig2}
    \vspace{-15pt}
\end{figure*}
The initial training stage, which involves learning the diffusion transformer with real driving data, enhances both the realism and diversity of the model. The subsequent training stage further improves the model's realism and controllability. MuDi-Pro evaluated on the real-world nuScence driving dataset \cite{nuscenes}, significantly enhances the effectiveness of guided sampling, while maintaining realism and diversity. This provides a novel approach for study of traffic scenario generation. To summarize, our contributions are as follows:

\begin{itemize}
\item Developing scene level diffusion transformer to ensure realistic traffic scenario generation.
\item Introducing a learning method similar to multi-task learning to allow one model to reinforce multiple guidance.
\item Fine-tuning the data-driven model through DPO including guided sampling, effectively addressing the trade-off between realism and controllability.
\end{itemize}

\section{Related Works}
\noindent
{\bf Traffic Simulation.} There are two principal approaches to traffic simulation methods: rule-based and learning-based. Rule-based approaches~\cite{rulebase1, rulebase2} analyze vehicle movements and control them along fixed paths. While these methods are intuitively understandable to users, they often lack the expressiveness necessary to accurately replicate real-world driver behavior, resulting in movements that may significantly deviate from actual driving patterns. In contrast, learning-based approaches~\cite{learningbase1, learningbase2, learningbase3, trace} use real-world traffic data~\cite{nuscenes} to train deep generative models such as VAE~\cite{vae} and diffusion model~\cite{diffusion}. Notably, recent advancements in diffusion models, which have demonstrated high performance, show great promise in traffic simulation by enabling the generation of highly realistic scenarios~\cite{motiondiffuser, traffcdiffusion1,trafficdiffusion2, trafficdiffusion3}. However, a notable limitation of learning-based approaches is their lack of controllability. In response, recent research have shown that it is possible to achieve controlled trajectory generation in learning-based models. Diffuser~\cite{diffuser} demonstrates great progress in this area, suggesting that trajectories generated by learning-based models can indeed be controlled. Recent works have sought to merge the strengths of learning-based and rule-based approaches to enhance both realism and controllability. For example, DiffScene~\cite{trafficdiffusion4} combines a diffusion model with adversarial optimization, while KING~\cite{king} use imitation learning with vehicle dynamics to achieve desired trajectories. Another approach, DJINN \cite{navigate} ensures controllability in a learning-based model by using a task-mask to control the trajectories generated, BehaviorGPT~\cite{behaviorgpt} and  InteractTraj~\cite{xia2024language} aimed to generate realistic interactive traffic trajectories using LLM. While these efforts concentrate on increasing realism and controllability of traffic scenarios, our work additionally focuses on scenario diversity, addressing a critical aspect of traffic simulation.

\noindent
{\bf Fine-tuning Diffusion Model.} Large-scale generative models, including diffusion models, have the potential to produce a broad spectrum of outcomes. Customizing these models to align with particular datasets or desired objectives is pivotal in the field of generative model research. Recent works have customized diffusion models by fine-tuning various components, such as the weights~\cite{fine-tuning1}, the embedding layer~\cite{fine-tuning2} or adaptors~\cite{fine-tuning3}, to better serve specific datasets. Additionally, other works~\cite{fine-tuning4,fine-tuning5} have focused on fine-tuning diffusion models for few-shot adaptation. Fine-tuning diffusion models significantly expands their capabilities, enabling them to address a wider range of tasks and preferences. Recent research~\cite{multitaskdiffusionfinetuning} has demonstrated that by applying multi-task learning during fine-tuning, a diffusion model can be adapted to cover multiple tasks simultaneously. Furthermore, other studies~\cite{diffusionDPO, diffusionrlhf, uehara2024feedback} use a fine-tuning approach to incorporate human preferences into the model. With the advent of Reinforcement Learning with Human Feedback~(RLHF) and DPO, methods for fine-tuning generative models to align with human preference, enable the incorporation of diverse intentions into models in terms of {\it preference}.

Our work builds upon these advancements, particularly inspired by DPO-SDXL~\cite{diffusionDPO}. We apply DPO, which is posited to be more effective than RLHF in certain contexts, as it bypasses the need for learning a reward model and consequently avoids the pitfalls of reward hacking~\cite{dpo}. By leveraging DPO, we aim to generate traffic scenarios that are not only more realistic and diverse but also more controllable, addressing key challenges in traffic simulation.


\section{Method} 

\subsection{Problem Formulation}

\noindent\textbf{Problem Formulation.} Building on previous research~\cite{ctg}, traffic scenario generation is defined as the challenge of developing and refining a model that guarantees realism, diversity, and controllability in traffic scenarios. Let the state at a time step $t$ be $s_t = [x_t, y_t, \theta_t, v_t, \dot{\theta}_t]$, including 2D location, heading, speed, and yaw rate. Similarly, let the action at a time step $t$, so-called as control, be $a_t = [ \dot{v}_t, \ddot{\theta}_t]$ with acceleration and yaw acceleration. Assuming a target, $tgt$ and $N$ neighbor agents in the traffic scenario at a time step $t$,  we introduce a map feature ${\textit M}^i = \epsilon_M(\text{M}^i)$ for each agent $i$, extracted by a map encoder ${\epsilon}_M(\cdot)$ from a local semantic map $\text{M}^i$. Further, past and future trajectory features are defined as ${\textit P}^i= \epsilon_P ([ s^i_{t-H_P}, \ldots, s^i_{t-1} ])$ and $ {\textit F}^i= \epsilon_F ( [ s^i_{t+1}, \ldots, s^i_{t+H_F} ])$ for each agent $i$, with the time horizons $H_j$ for $j \in \{P,F\}$, respectively, extracted by trajectory encoder ${\epsilon}_j(\cdot)$ for $j \in \{P,F\}$ . These extracted features collectively define a scene context as $ {\textit C} = ({ \textit M}, {\textit P}, {\textit F})$, where ${\textit M} = \{ {\textit M}^{tgt},{\textit M}^{1},\ldots,{\textit M}^{N} \}$, ${\textit P} = \{ {\textit P}^{tgt},{\textit P}^{1},\ldots,{\textit P}^{N} \}$, and ${\textit F} = \{ {\textit F}^{tgt},{\textit F}^{1},\ldots,{\textit F}^{N} \}$ for all agents.
We also denote the future action trajectory as $T_a := \begin{bmatrix} a_0 & \ldots & a_{H_F-1} \end{bmatrix} $ and the (future state) trajectory as $T_s := \begin{bmatrix} s_1 & \ldots & s_{H_F} \end{bmatrix} $, respectively, over the forthcoming $H_F$ time steps. Our goal is to generate a new, diverse trajectory $T_s$ that exhibits both realistic and rule-compliant traffic behavior. To this end, we apply a transition function $f$, which computes $s_{t+1} = f(s_t, a_t)$ given the state $s_{t}$ and the action $a_t$, based on a unicycle dynamics model for simple vehicle dynamics~\cite{luca1988feedback}. By  predicting $T_a$ and obtaining $T_s$ through rollout from the initial state $s_0$ using $f$, we ensure that physical feasibility of the state trajectory emanating from the denoising process. 

\subsection{Training Diffusion Model for Traffic Scenario Generation}

\noindent\textbf{Trajectory Diffusion Formulation.} The training of a diffusion model involves developing a denoising process, which functions inversely to the forward diffusion process. The forward process incrementally introduces noise to a clean trajectory over $K$ diffusion steps until it becomes completely noisy. Let $T_s^k$ represent the action trajectory at the $k$-th diffusion step. Starting from the original clean trajectory $T_s^0$, the forward diffusion process is defined as:
\vspace{-5pt}
\begin{equation}
\label{eq:forward_process}
\begin{split}
q(T_s^{1:K} | T_s^0) &:= \prod_{k=1}^K q(T_s^k | T_s^{k-1}) \\
 &:= \prod_{k=1}^K \mathcal{N}(T_s^k ; \sqrt{1 - \beta_k}T_s^{k-1}, \beta_k I)
\end{split}
\end{equation}
where each $\beta_k$ for $k=1,2,\ldots,K$, specifies a predetermined variance schedule that controls the level of noise added at each diffusion step $k$. With a sufficiently large $K$, we approximate $q(T_s^K) \approx \mathcal{N}(T_s^K; 0, I)$.
To reverse the diffusion process, we utilize the Improved Denoising Diffusion Probabilistic Models methodology~\cite{improved}, similar to the TRACE framework~\cite{trace}. This approach is particularly effective for guided sampling, as it directly yields clean trajectories. The traffic model is designed to learn the inverse process, transforming sampled noise back into plausible trajectories. Each iteration of this reverse process is explicitly conditioned on the scene context $\textit C$, ensuring that the output remains consistent with predefined constraints. Given the scene context $\textit C$, the reverse process is formulated as follows:
\begin{equation}
\label{eq:reverse_process}
    p_\theta(T_s^{k-1}|T_s^{k}, C) := \mathcal{N}(T_s^{k-1}; \mu_\theta(T_s^{k}, k, C), \Sigma_k(T_s^{k}, k, C)) \\
\end{equation}
\noindent
where\(\ \mu_\theta(T_s^{k}, k, C)\) and\(\ \Sigma_k(T_s^{k}, k, C) \) are the mean and variance of the reverse process at diffusion step \( k\), respectively, with \(\theta\) representing the model parameters. Note that $T_s^{k-1}$, the model output, is generated from $T_a^{k-1}$, which produces a noisy trajectory using known dynamics. During the denoising phase, the traffic model learns to parameterize the mean of the Gaussian distribution at each diffusion step $k$. 

\noindent
{\bf Classifier-free Sampling\footnote{we refer to it as “sampling” instead of the common term “guidance” to avoid confusion with the guidance of guided sampling.} and Clean Trajectory Guided Sampling.} 
For guided sampling, our model predicts a clean trajectory \( T_s^0\) at each diffusion step $k$, following the TRACE framework \cite{trace}. To further enhance diversity, we implement a strategy that combines Classifier-Free Sampling~(CFS) with clean trajectory-guided sampling during test time.\\
\noindent
For CFS, we simultaneously train a future-conditioned model \( p_{\theta}(T_s^k,k,{ \textit M},{\textit P},{\textit F})\) and a non-future-conditioned model \( p_{\theta}(T_s^k,k,{\textit M},{\textit P})\) utilizing future feature condition dropout. This approach enables the model to generate trajectories that reflect future conditions with varying degrees at test time. The trajectories predicted by both models are merged using a CFS weight $w$, as follows:
%
%
\begin{equation}
    T_s^0 = w p_{\theta}(T_s^k,k,{\textit M},{\textit P},{\textit F}) + (1-w) p_{\theta}(T_s^k,k,{\textit M},{\textit P}).\\
\end{equation}
\noindent
Here,  \( T_s^0\) represents the clean trajectory from model \( p_{\theta}\), modulated by CFS and controlled by $w$. Setting  \( w=1.0\) fully incorporates future information, whereas \( w=0.0\) generates trajectories without considering future information. This mechanism ensures that the generated trajectory adheres to the guided sampling principles. 
Previous works such as Diffuser \cite{diffuser}, directly perturb the noisy mean predicted by the network to align with guidance. However, this approach, reliant on a learned loss function, necessitates training across a spectrum of noise levels and encounters numerical challenges with analytical loss functions. To circumvent these issues, Trace \cite{trace} employs a {\it reconstruction guidance} strategy \cite{video-diffusion-model}, extending the guidance formulation to work with any arbitrary guidance function. Following the approach of Trace, we perturb the clean trajectory as:
\begin{equation}
    \tilde{T}_s^0 = T_s^0-\alpha\Sigma_k\nabla_{T_s^{k}}\mathcal{J}(T_s^0)
\end{equation}
\noindent
where\(\ \alpha\) denotes the guidance strength, \( \mathcal{J} \) denotes selected specific guidance. The noisy mean \( \tilde{\mu}_\theta \) is then calculated as described in Eq.~(2), treating \( \tilde{T}_s^0 \) as the network's output. This process of generating a clean trajectory is executed at each diffusion step, ensuring robust controllability, as illustrated in \Cref{figure2-b}.

\noindent
{\bf Architecture: Traffic Diffusion Transformer.} We present the Diffusion Transformer (DiT), which is tailored for generating traffic scenario, inspired by the Vision Transformer (ViT). This section elaborates on the forward pass of the DiT. \Cref{figure2-a} provides an overview of the DiT architecture, illustrating its design. The architecture of the DiT is structured around three core components: encoders for tokenizing input data, transformer blocks for processing the tokenized noisy input data along with scene context $\textit C$, and a transformer decoder aimed at reconstructing the data to its original format.
Noisy trajectory and traffic information are tokenized using an MLP encoder, producing the trajectory tokens  \( T_{\tau}\in\mathbb{R}^{6\times D_H}\) and scene context tokens \( \textit C\in\mathbb{R}^{D_H}\), where \( D_H\) denotes the hidden dimension. These tokens are then processed by the Transformer block. Building on prior research~\cite{ditrefer, dit} that highlights the advantages of initializing residual blocks as identity functions for enhancing training efficiency  in large-scale training, we adapt this approach for our Diffusion U-Net models. 
Building upon the methodologies of previous studies~\cite{dit}, our model further utilizes context information to adjust not only the parameters $\gamma$ and $\beta$ but also introduces dimension-wise scaling parameters $\alpha$, which are incorporated immediately prior to the residual connections in the model's architecture. Following the final Transformer block, the transformer decoder translates the context tokens into a clean action. This clean action is subsequently refined through the known dynamics to produce a clean trajectory.
\begin{figure}[t]
    \vspace{2mm}
    \centering
    \begin{subfigure}[c]{.48\columnwidth}
        \centering
        \includegraphics[width=\textwidth]{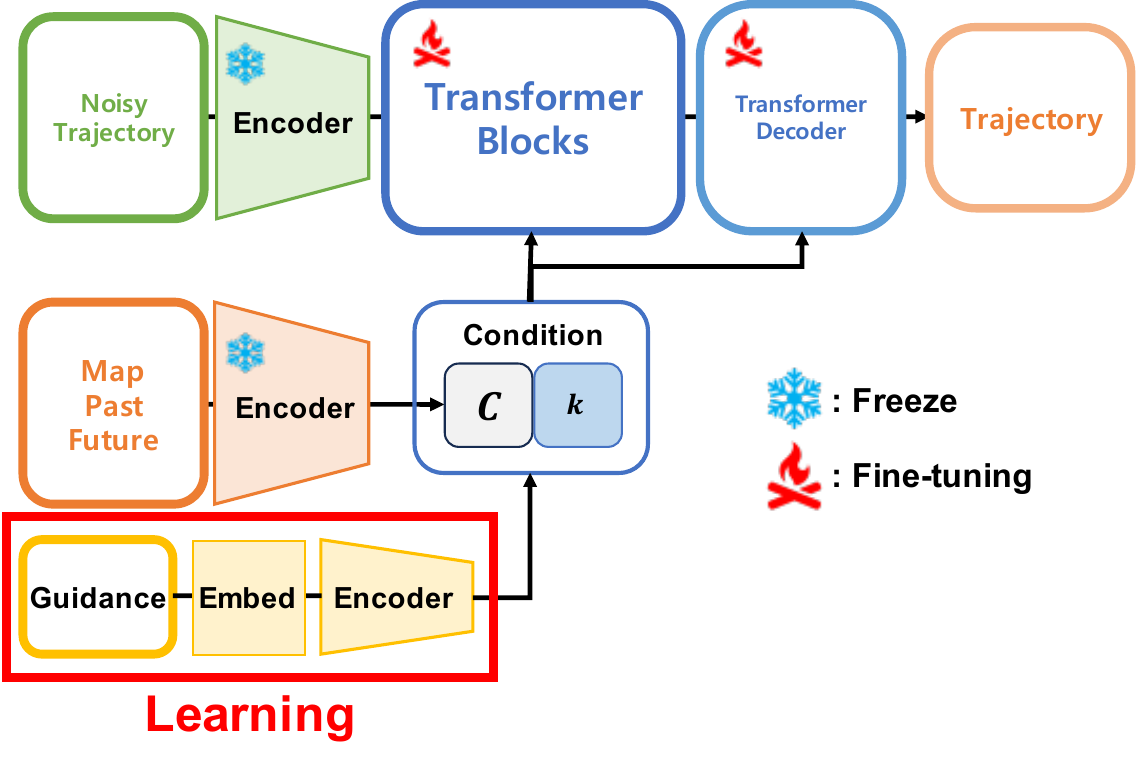}
        \caption{Guidance Conditional Layer}
        \label{figure3-a}
    \end{subfigure}
    ~ 
    \begin{subfigure}[c]{.48\columnwidth}
        \centering
        \includegraphics[width=\textwidth]{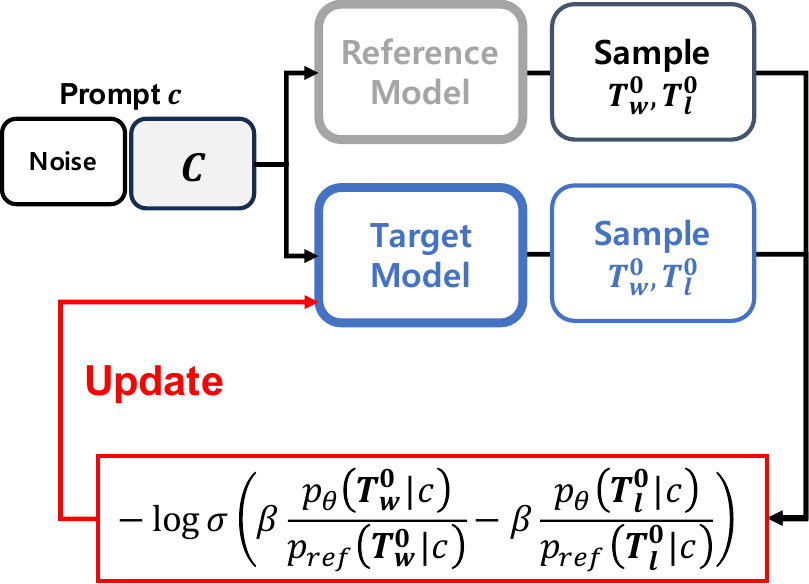}
        \caption{DPO Fine-tuning}
        \label{figure3-b}
    \end{subfigure}
    \caption{{\bf Fine-tuning Framework.} Fine-tuning is conducted in two phases: (a) The guidance conditional layer fine-tunes a segment of the transformer based on chosen guidance. (b) DPO further fine-tunes the target model using DPO loss.}
    \label{fig3}
    \vspace{-10pt}
\end{figure}

{\centering
\begin{figure}[t]
\vspace{-2mm}
\begin{minipage}{\linewidth}
  \begin{algorithm}[H]
  \footnotesize
    \caption{Fine-tuning with DPO}
    \label{fine-tuning}
    \begin{algorithmic}[1]
    \State {{\bf Require:} pre-trained diffusion model $\epsilon_{\theta}$, prepared dataset ${\mathcal{D}}$,  DPO scale $\beta$, fine-tuning epoch $N$}
    \State Copy pre-trained model $\epsilon_{ref} = \epsilon_{\theta}$ 
    \State Set $\epsilon_{ref}$ to untrainable
    \While{not converge}
        \State Choose prompt $c$, winning and losing samples $(T_{w}, T_l)$ from $\mathcal{D}$
        \For{$n = 1, \dots, N$}
            \State $\#$ Guided Sampling
       \State Sample winning sample $T_{w,\theta}$, lossing sample $T_{l,\theta}$ from $\epsilon_{\theta}(c)$
       \State Sample winning sample $T_{w,ref}$, lossing sample $T_{l,ref}$ from $\epsilon_{ref}(c)$
       \State Compute error $E_{i,j} = \text{L}_2(T_i,T_{i,j})$; $i\sim [{\it w, l}], j\sim[\theta,{\it ref}]$
       \State Compute different $d_i = E_{i,\theta} - E_{i,{\it ref}}$; $i\sim [{\it w, l}]$
       \State $\mathcal{L}_{\text{DPO}} = -\log \sigma \left( -\beta(d_{\it w}-d_{\it l}) \right)$
       \State Take gradient step on $\nabla_{\theta}\mathcal{L}_{\text{DPO}}$
        \EndFor
    \EndWhile
    \end{algorithmic}
  \end{algorithm}
\end{minipage}
\vspace{-6mm}
\end{figure}
}
\subsection{Fine-tuning Diffusion Model}
\noindent
This section outlines the process for fine-tuning in two steps. Initially, the model is fine-tuned using a guidance conditional layer, leading to the development of MuDi, which applies a multi-task learning approach. Subsequently, by further fine-tuning the model with DPO, MuDi-Pro is developed.

\noindent
{\bf Guidance Conditional Layer.} Drawing on recent advancements in task conditional approaches and multi-task diffusion fine-tuning~\cite{taskconditional, multitaskdiffusionfinetuning}, we introduce a novel network architecture featuring a guidance conditional layer. \Cref{figure3-a} illustrates this framework, which selects specific guidance \( \mathcal{J} \). As shown in the red box in the \Cref{figure3-a}, the architecture of the guidance conditional layer is bifurcated into two main components: a guidance embedding layer and a guidance encoding network. The former translates the provided guidance into a distinct vector \( v_{\mathcal{J}}\in \mathbb{R}^{D_{H/4}} \), whereas the latter converts this vector \( v_{\mathcal{J}} \) into a unique guidance-latent space \( l_{\mathcal{J}}\in \mathbb{R}^{D_{H}} \), where \( D_{H/4} \) represents a quarter of the hidden dimension size \( D_{H} \). This guidance-latent is then merged with scene context for processing in a transformer block. To optimize fine-tuning efficiency, modifications are restricted to the transformer blocks and the concluding layer. This strategy allows the transformer block to adapt its function based on the provided guidance, enabling a singular shared layer to act as a multi-decoder, a common feature in multi-task learning.
\begin{figure}[t!]
  \vspace{1mm}
  \centering
  \includegraphics[width=\linewidth]{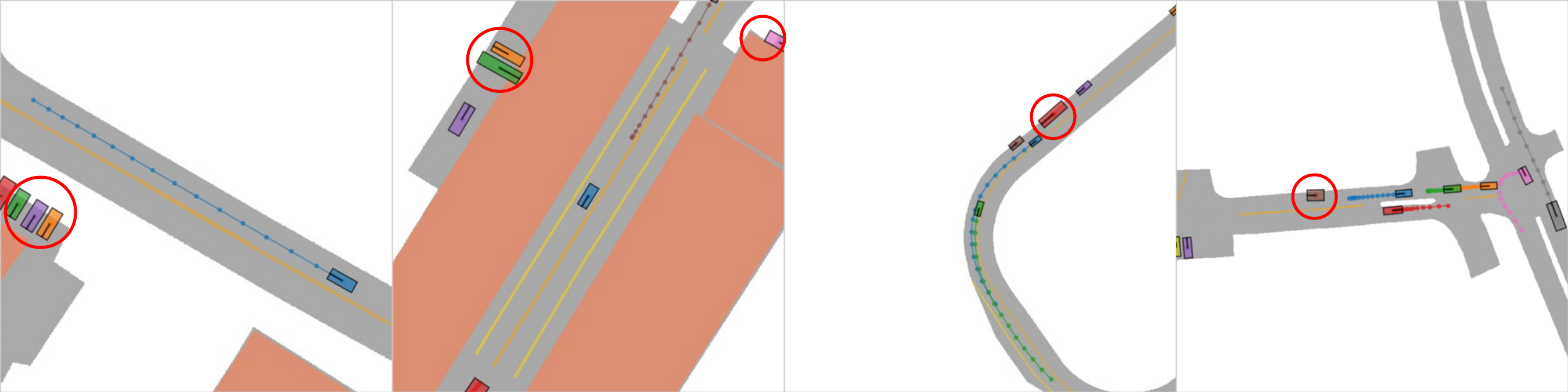}
  \caption{Examples of sample data exhibiting collisions from the onset of traffic, with collision points highlighted by \textcolor{red}{red circles}. Most of the vehicles with collisions are parked and have stopped slightly off roads.} 
  \label{collGT}
  \vspace{-2mm}
\end{figure}

\begin{table}[t!]
  \centering
\caption{Backbone Model Comparison.}
\vspace{-8pt}
\label{tab1}
  \begin{tabular}{l|cc|cc|c}
 \multicolumn{6}{r}{\tiny * the best results are highlighted.}\\
    \toprule
    &  \multicolumn{2}{c|}{{\bf Data-driven}} & \multicolumn{2}{c|}{{\bf Stability}} & \multicolumn{1}{c}{\bf Realism} \\
    & pos& ang& map& veh& real\\
    & (m)& (deg)& (\%)& (\%)& ($10^{-3}$)\\
    \midrule
    STRIVE(recon)  & 0.7621 & 1.619 & 2.67 & 4.71 & 7.1\\
    STRIVE(sample)  & 0.8796 & 1.682 & 5.39 & 5.98 & 7.3\\
    CTG &  0.6514 & 1.046 & 2.31 & 4.96 & 3.7\\
    CTG++ & 0.5794 & 0.9306 & 3.92 & 4.54 & 4.4\\
    \midrule
    Ours(recon) & {\bf 0.5080} & {\bf 0.7749} & {\bf 0.79} & {\bf 1.19} & {\bf 1.9}\\
    Ours(mix) & 1.487 & 1.626 & 2.27 & 4.14 & {\bf 1.9}\\
    Ours(sample) & 2.738 & 2.682 & 4.84 & 7.97 & 2.0\\
    \bottomrule
  \end{tabular}
  \vspace{-17pt}
\end{table}
\noindent
{\bf Fine-tuning via DPO.} Contrary to previous studies~\cite{d3po,diffusionDPO} that apply DPO to capture human preferences, our approach leverages DPO to refine our model as shown in the \Cref{figure3-b}, thereby enhancing the efficiency of guided sampling. We devise a preference dataset predicated on the guidance score, which is directly derived from the guidance loss during the diffusion model's sampling phase, to provide heuristic-based preferences for intuitiveness, similar to recent work~\cite{du2024exploration}. Within the DPO framework, we employ the prompt $c$, encompassing all traffic information and Gaussian noise. Subsequently, we compile the DPO dataset \( \mathcal{D} = \{ (c, T^0_w, T^0_l) \}\). 
\begin{figure*}[t]
    \vspace{2mm}
    \centering
    \begin{subfigure}{.32\textwidth}
        \centering
        \includegraphics[height=5.6cm]{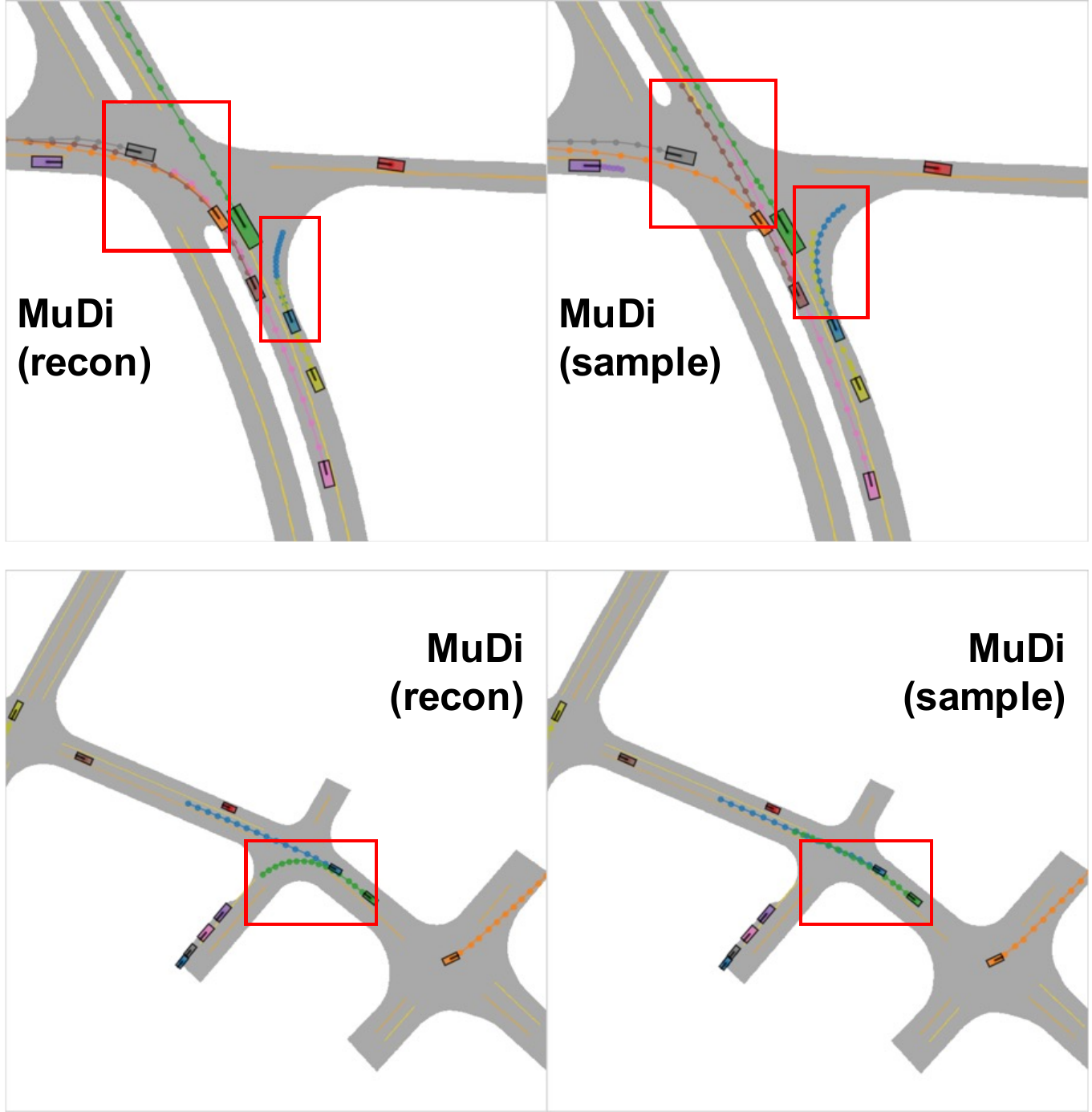}
        \caption{MuDi (recon vs sample)}
        \label{figure5-a}
    \end{subfigure}
    \begin{subfigure}{.32\textwidth}
        \centering
        \includegraphics[height=5.6cm]{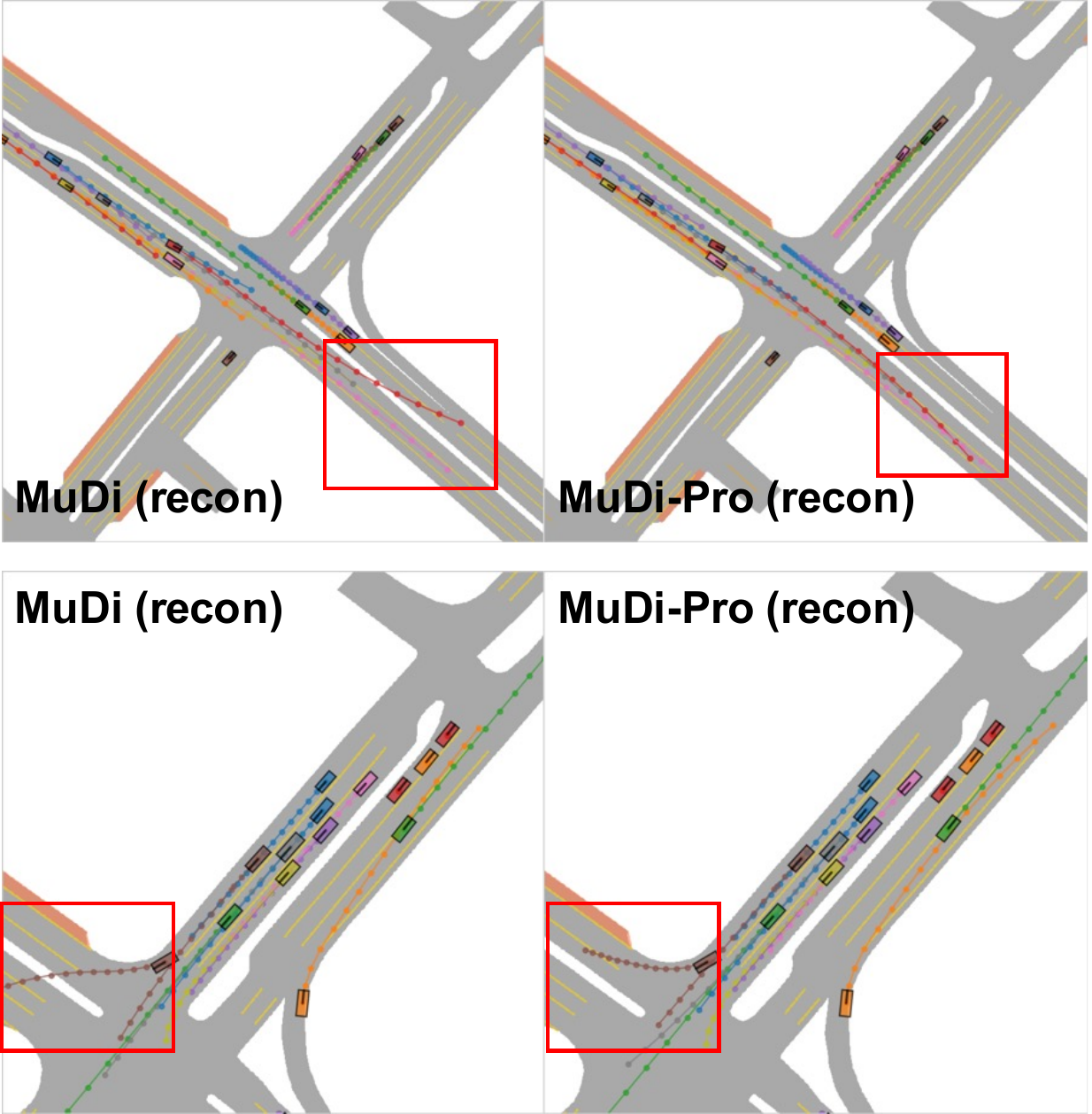}
        \caption{Mudi vs MuDi-Pro}
        \label{figure5-b}
    \end{subfigure}
    \begin{subfigure}{.32\textwidth}
        \centering
        \includegraphics[height=5.6cm]{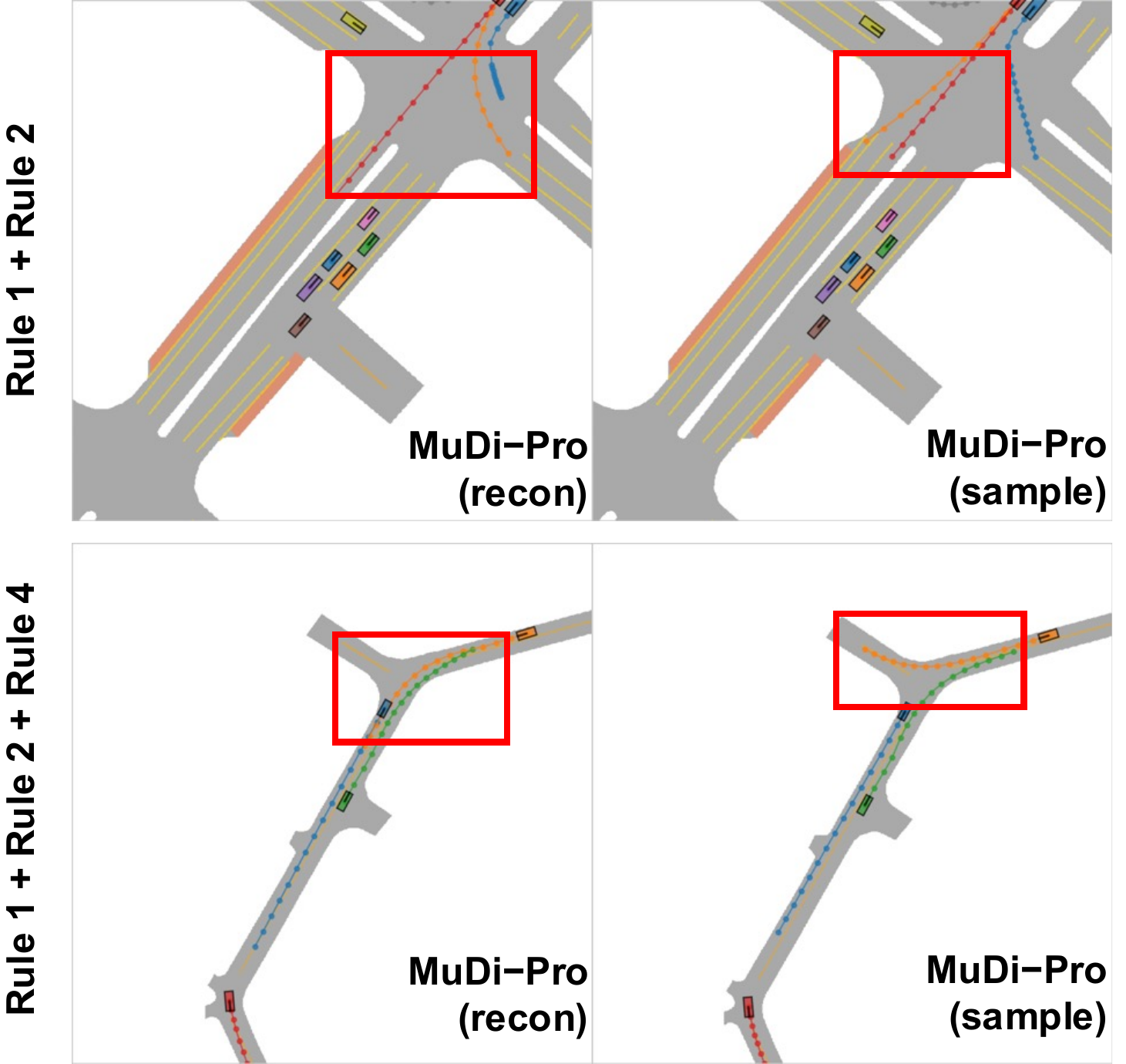}
        \caption{MuDi-Pro (recon vs sample)}
        \label{figure5-c}
    \end{subfigure}
    \caption{{\bf Qualitative results of MuDi and MuDi-Pro.} (a) Qualitative results demonstrate that our model can produce diverse samples with varying {\it w}. (b) Qualitative results of comparing MuDi-Pro with MuDi. MuDi-Pro produces plausible and realistic trajectories in scenes where MuDi fails to do so. (c) Qualitative diversity results of MuDi-Pro. The generated trajectories, from both the sample mode and the reconstruct mode, tends to exhibit multi-modal characteristics. Significant differences are highlighted with \textcolor{red}{red squares}.}
    \label{figall}
    \vspace{-5pt}
\end{figure*}

\begin{table*}[t]
  \caption{Controllability Evaluation. We evaluate five single-rule cases and five multiple-rule cases for several combinations.}
  \vspace{-8pt}
  \label{tab2}
  \centering
  \resizebox{\textwidth}{!}{%
    \begin{tabular}{l|cccc|cccc|cccc|cccc|cccc|cccc}
     \multicolumn{25}{r}{ * the best results are highlighted.}\\
      \toprule
       & \multicolumn{4}{c|}{{\bf rule 1}} & \multicolumn{4}{c|}{{\bf rule 2}} & \multicolumn{4}{c|}{{\bf rule 3}} & \multicolumn{4}{c|}{{\bf rule 4}} & \multicolumn{4}{c|}{{\bf rule 1+2}} & \multicolumn{4}{c}{{\bf rule 1+2+4}} \\
 & \multicolumn{4}{c|}{{\bf no vehicle collision}}& \multicolumn{4}{c|}{{\bf no off-road}}& \multicolumn{4}{c|}{{\bf goal waypoint}}& \multicolumn{4}{c|}{{\bf target speed}}& \multicolumn{4}{c|}{{\bf multiple rule}} & \multicolumn{4}{c}{{\bf multiple rule}}\\
        & map & veh & rule & real & map & veh & rule & real & map & veh & rule & real & map & veh & rule & real & map & veh & rule & real & map & veh & rule & real \\ 
        & (\%) & (\%) & ($10^{-3}$) & ($10^{-3}$) & (\%) & (\%) & ($10^{-3}$) & ($10^{-3}$) & (\%) & (\%) & ($10^{-3}$) & ($10^{-3}$) & (\%) & (\%) & ($10^{-3}$) & ($10^{-3}$) & (\%) & (\%) & ($10^{-3}$) & ($10^{-3}$) & (\%) & (\%) & ($10^{-3}$) & ($10^{-3}$)\\ 
       \midrule
       STRIVE(recon)+opt & {\bf 5.10}& 0.61& 9.3& 7.6& {\bf 1.84}& 3.13& 46.4& 7.3& 4.76& 2.31& 168.2& 7.0& 5.23 & {\bf 1.90} & 110.6 & 7.2 & {\bf 1.97} & 1.02 & 57.9 & 6.3 & {\bf 1.98} & 0.94 & 208.9 & 7.0 \\
       STRIVE(sample)+opt & {\bf 5.10} & 1.29 & 25.1 & 6.9 & 1.97& 5.17 & 47.8 & 6.9 & {\bf 4.49} & 3.94 & 173.0 & 6.8 & {\bf 4.22} & 3.61 & 116.5 & 6.8 & 2.24 & 2.65 & 82.9 & 6.9 & 2.58 & 2.38 & 202.5 & 6.5 \\
       CTG & 8.92 & 0.79 & 15.8 & 4.9& {\bf 1.84} & 9.68 & 46.4 & 4.6 & 5.10 & 3.54 & 169.8 & 3.6 & 5.23 & 3.13 & 103.4 & 5.5 & {\bf 1.84} & 4.65 & 70.5 & 3.7 & 2.37 & 4.29 & 196.3 & 3.6\\
       CTG++ & 8.92 & 0.61 & 9.3 & 4.1 & 2.52 & 5.45 & 54.1 & 4.2 & 6.12 & 3.13 & 167.2 & 4.1 & {\bf 4.22} & 2.87 & 97.5 & 5.2 &  3.41 & 4.29 & 90.5 & 5.1 & 2.58 & 3.92 & 162.7 & 4.2\\
       \midrule
       MuDi(recon) & 7.19 & {\bf 0.21} & {\bf 2.1} & {\bf 2.0} & 2.01 & {\bf 1.36} & {\bf 45.5} & {\bf 2.1} & 6.32 & {\bf 0.85} & {\bf 165.6} & 3.4 & 11.76 & 2.78 & 92.8 & 3.6 & 2.12 & {\bf 0.20} & 52.4 & {\bf 2.1} & 2.14 & {\bf 0.19} & 152.1 & {\bf 3.4}\\
       MuDi(mix) & 7.19 & {\bf 0.21} & 2.2 & {\bf 2.0} & 2.02 & {\bf 1.36} & 45.8 & {\bf 2.1} & 6.32 & {\bf 0.85} & 165.7 & 3.4 & 11.77 & 2.87 & 92.8 & 3.6 & 2.12 & {\bf 0.20} & 52.4 & {\bf 2.1} & 2.11 & 0.20 & {\bf 151.5} & {\bf 3.4}\\
       MuDi(sample) & 11.19 & 2.48 & 24.1 & 2.1 & 2.81 & 8.26 & 52.9 & 2.2 & 8.92 & 6.07 & 165.7 & {\bf 3.2} & 13.32 & 8.11 & {\bf 89.6} & {\bf 3.5} & 3.39 & 2.76 & {\bf 52.3} & 2.3 & 2.14 & 0.20 & 153.5 & {\bf 3.4}\\ 
       \bottomrule
    \end{tabular}%
    }
    \vspace{-15pt}
\end{table*}

This dataset includes instances designated as the \emph{winning} sample \(\ T^0_w\) and the \emph{losing} sample\(\ T^0_l\), both derived from the same prompt $c$. From the generated pair, the sample with the higher preference score is identified as the \emph{winning} sample, whereas its counterpart with the lesser preference score is labeled the \emph{losing} sample. Following the dataset preparation, we fine-tuned our model using DPO loss as follows:

{\small
\begin{equation}
L_{DPO}(\theta) = -\mathbb{E}_{c, T_w^0, T_l^0} \left[ \log \sigma \left( \beta \log \frac{p_\theta(T_w^0 | c) \cdot p_{\text{ref}}(T_l^0 | c)}{p_{\text{ref}}(T_w^0 | c) \cdot p_\theta(T_l^0 | c)} \right) \right]
\end{equation}
}
\noindent
where, \( \sigma\) denotes the sigmoid function, while \( \beta\) serves as a hyperparameter to govern regularization. The model \(p_\theta\) is designated for fine-tuning, and \( p_{\text{ref}}\) represents the reference model, which remains unchanged during this process. The model is reparameterized to enable direct optimization towards the preferred distribution \( p_\theta(T_w^0 | c)\). The DPO process is executed over $\eta$ epochs, enhancing the model's capability to produce trajectories that are both realistic and compliant with established rules. Further details on the fine-tuning process are provided in Algorithm \Cref{fine-tuning}.


\section{Experiments}
We conduct experiments to show (1) our backbone model, MuDi generates diverse and realistic traffic scenarios, and (2) when compared, the fine-tuned model exhibits superior performance to strong baselines in traffic scenario generation.

\subsection{Experimental Setup}
\label{setup}
\noindent
\textbf{Dataset.} The nuScenes Dataset~\cite{nuscenes} is one of the large-scale real world driving datasets. It comprises total 1000 scenes, featuring 5.5 hours of precise travel paths in two urban areas with varying scenarios and congested traffic conditions. In using the nuScenes dataset, we encounter challenges that make the dataset difficult to utilize without prior dataset pre-processing. The dataset includes data that can negatively impact the learning process or render the evaluation results unclear. For example, as shown in \Cref{collGT}, several agents have polygons not fully included in the drivable area of the binary map. In training subset, 13.9\% of agents already have collisions in their past trajectories. This induces unintended map collision loss from the trajectories of these parked or stopped agents, causing trained traffic models to generate meaningless movements by accelerating to change positions in an attempt to avoid collisions that have already occurred. Our observations, along with a detailed result analysis, present the necessity of excluding cases where an agent's past trajectory indicates a collision from the loss function calculation. Moreover, such data are also excluded from the evaluation results of both baseline and our model.

\noindent
\textbf{Metrics. }Following \cite{ctg,metricsrefer,strive}, we evaluate the generated trajectories based on stability, controllability, and realism. {\it Stability} is measured by off-road and inter-vehicle collisions, represented as percentages: map for off-road collisions and veh for vehicle collisions. {\it Controllability} is assessed by how well the trajectories follow guidance, measured by the final guided loss (\textbf{rule}). {\it Realism} is evaluated by comfort, averaging forward/lateral accelerations and jerk (\textbf{real}). Additionally, the {\it data-driven capability} is assessed by comparing the generated trajectories to the dataset in terms of 2D position (\textbf{pos}) and heading angle (\textbf{ang}). Lower values for map, veh, rule, and real indicate better performance, while for pos and ang, lower values suggest better data alignment, and higher values indicate more trajectory diversity.

\noindent
\textbf{Baseline.} We compare our model to several baselines, including STRIVE~\cite{strive}, CTG~\cite{ctg}, and CTG++~\cite{ctg+llm}. STRIVE uses a VAE for traffic scenario generation with simple vehicle dynamics. We evaluate STRIVE in two modes: (1) sample mode (no future information) and (2) reconstruction mode (uses future information). The sample mode is comparable to our model at {\it w}=0.0, and the reconstruction mode at {\it w}=1.0. Additionally, we consider STRIVE+opt, which applies optimization for improved controllability. CTG and CTG++ are included as they use diffusion models for trajectory generation. All models are trained and evaluated on the nuScenes~\cite{nuscenes} dataset.
For our model, we provide results for three different setups: "Ours (recon)" with {\it w}=1.0, "Ours (mix)" with {\it w}=0.5, and "Ours (sample)" with {\it w}=0.0. These correspond to varying the weighting parameters within our model to evaluate performance across different conditions.

\noindent
{\bf Map Collision.} For a fair comparison with the baseline, similar to STRIVE, we only consider map collisions for the ego agent during the training process and consider all agents during the guided sampling process, detailed in the supplementary materials. Therefore, we evaluate map collisions for ego agent in \cref{sec4.2}, and for all agents in \cref{sec4.3,sec4.4}.

\subsection{Evaluation on Data-driven Ability}
\label{sec4.2}
Before evaluating controllability, we conduct a quantitative assessment of the model's ability to capture the data distribution. We compare our backbone model, which does not use guided sampling, with baseline. To this end, we significantly utilize the {\bf pos} and {\bf ang} metrics to access  the similarity between generated samples and the real sample. In evaluating data-driven ability, lower values of these metrics indicate better performance. \Cref{tab1} shows that ours~(recon) outperforms baselines with all metrics, and ours~(sample) also outperforms baselines with all metrics except for the {\bf veh} metric value. Furthermore, we demonstrate the diversity of samples generated by our model by varying the CFS weight \({\it w}\) as illustrated in \Cref{figure5-a}. Assuming that reconstruction results align with the ground truth data, the sampled trajectories show well a range of possible outcomes.

\subsection{Evaluation on Controllability }
\label{sec4.3}
In this section, we evaluate the controllability of MuDi, which hasn't been fine-tuned, compared to baselines. We report the stability, controllability, and realism of both models in \Cref{tab2}. As shown in \Cref{tab2}, we apply four single-rule cases and also conduct experiments with three multiple-rule cases by combining them. In many cases, MuDi outperforms the baselines by registering lower scores than all baselines.

\subsection{Fine-tuned Model Evaluation}
\label{sec4.4}
Finally, we evaluate our refined model, MuDi-Pro, which is a fine-tuned version of MuDi with an added  guidance conditional layer and DPO. To assess the impact of fine-tuning, we compare MuDi-Pro to MuDi, a strong backbone model. As indicated in \Cref{tab3}, MuDi-Pro~(recon) outperforms MuDi~(recon) in nearly all metrics across various guidance combinations. Furthermore, even MuDi-Pro~(sample) also outperforms MuDi~(recon) in almost all metrics while maintaining diversity. Higher values of {\bf pos} and {\bf ang} indicate better diversity; therefore these values are highlighted. The qualitative differences are evident in \Cref{figure5-b}, where the off-road occurrences are not occurred even \textit{rule 1} is given as guidance in MuDi~(recon), yet such occurrences are absent in the MuDi-Pro~(recon). By examining the recon scores in \Cref{tab3} and the samples in \Cref{figure5-c}, it is clear that the generated samples remain diverse, despite the increased controllability in guided sampling.

\begin{table}[t]
  \vspace{1mm}
  \caption{Fine-tuned Model Evaluation.}
  \vspace{-8pt}
  \label{tab3}
  \centering
  \resizebox{\columnwidth}{!}{%
    \begin{tabular}{l|cccccc}
     \multicolumn{7}{r}{\tiny * the best results are highlighted.}\\
      \toprule
       &  \multicolumn{6}{c}{{\bf rule 1+2}} \\
       &    pos& ang& map& veh& rule & real \\
 & (m)& (deg)& (\%)& (\%)& ($10^{-3}$)& ($10^{-3}$)\\ 
       \midrule
       MuDi(recon) & 0.5175 & 0.8082 & 2.12& 0.2& 52.4& {\bf2.1}\\
       MuDi-Pro(recon) & 0.4186 & 0.7618 & {\bf1.73}& {\bf0.03}& {\bf43.5}& {\bf2.1}\\
       MuDi-Pro(sample) & {\bf 2.7200}& {\bf 2.9233} & 1.77& {\bf0.03}& 43.6& {\bf2.1}\\  
       \midrule
       \midrule
       &  \multicolumn{6}{c}{{\bf rule 1+2+4}}  \\
       &    pos& ang& map& veh& rule & real \\
       \midrule
       MuDi(recon) &  0.9954 & 1.1678 & 2.14& 0.19& 152.1& {\bf3.4}\\
       MuDi-Pro(recon) &  0.9385 & 1.0976 & {\bf1.75}& 0.03& 138& 3.8\\
       MuDi-Pro(sample) & {\bf 2.9044} & {\bf 3.1537} &  1.76& {\bf0.02}& {\bf137.8}& 3.8\\  
       \bottomrule
    \end{tabular}%
    }
    \vspace{-8pt}
\end{table}

\subsection{Ablation Study}
Ablation studies in Tab.~\ref{tab4} highlight the impact of key components in our model. We evaluate MuDi and its variants without the unicycle model, transformer, data filtering, and past/future context. Additionally, we investigate the performance of MuDi-Pro variants, focusing on the contributions of the decoder, transformer blocks, and the full model (denoted as Ours). These results validate the significance of each component in enhancing the overall performance.

\begin{table}[t!]
  \centering
\caption{Ablation Study.}
\vspace{-8pt}
\label{tab4}
  \begin{tabular}{l|cccc}
 \multicolumn{5}{r}{\tiny * the best results are highlighted.}\\
    \toprule
    & map& veh & rule & real\\
    & (\%) & (\%) & ($10^{-3}$) & ($10^{-3}$)\\
    \midrule
    MuDi & 2.21 & 0.20 & 52.4 & 2.1\\
    MuDi(-unicycle) & 13.81 & 10.7 & 361.2 & 8.1\\
    MuDi(-transformer) & 4.96 & 1.54 & 56.7 & 5.8\\
    MuDi (-data filtering) & 2.42 & 2.24 & 73.1 & 3.8\\
    MuDi (-past\&future)& 8.33 & 2.06 & 127.3 & 4.9\\
    \midrule
    MuDi-Pro(decoder) & 2.06 & 0.19 & 51.7 & 2.1\\
    MuDi-Pro(blocks) & 1.89 & 0.12 & 48.1 & 2.1\\
    MuDi-Pro(Ours) & {\bf1.73}& {\bf0.03}& {\bf43.5}& {\bf2.1}\\
    \bottomrule
  \end{tabular}
  \vspace{-15pt}
\end{table}

\section{Conclusion}
We introduced MuDi-Pro, a multi-guided diffusion model using direct preference optimization to generate realistic, diverse, and controllable traffic scenarios. It effectively balances realism and controllability, but its classifier-free sampling is limited to trajectory space. Future work will explore latent space sampling and improve the efficiency of the guided fine-tuning process to extend its capabilities.



\clearpage

\bibliographystyle{unsrt}
\bibliography{IEEEfull}

\begin{thebibliography}{10}

\bibitem{review}
Di~Chen, Meixin Zhu, Hao Yang, Xuesong Wang, and Yinhai Wang.
\newblock Data-driven traffic simulation: A comprehensive review.
\newblock {\em arXiv preprint arXiv:2310.15975}, 2023.

\bibitem{intro1}
Nidhi Kalra and Susan~M Paddock.
\newblock Driving to safety: How many miles of driving would it take to demonstrate autonomous vehicle reliability?
\newblock {\em Transportation Research Part A: Policy and Practice}, 94:182--193, 2016.

\bibitem{ctg+llm}
Ziyuan Zhong, Davis Rempe, Yuxiao Chen, Boris Ivanovic, Yulong Cao, Danfei Xu, Marco Pavone, and Baishakhi Ray.
\newblock Language-guided traffic simulation via scene-level diffusion.
\newblock {\em arXiv preprint arXiv:2306.06344}, 2023.

\bibitem{trafficRLHF}
Yulong Cao, Boris Ivanovic, Chaowei Xiao, and Marco Pavone.
\newblock Reinforcement learning with human feedback for realistic traffic simulation.
\newblock {\em arXiv preprint arXiv:2309.00709}, 2023.

\bibitem{realgen}
Wenhao Ding, Yulong Cao, Ding Zhao, Chaowei Xiao, and Marco Pavone.
\newblock Realgen: Retrieval augmented generation for controllable traffic scenarios.
\newblock {\em arXiv preprint arXiv:2312.13303}, 2023.

\bibitem{scenecontrol}
Jack Lu, Kelvin Wong, Chris Zhang, Simon Suo, and Raquel Urtasun.
\newblock Scenecontrol: Diffusion for controllable traffic scene generation.
\newblock In {\em 2024 IEEE International Conference on Robotics and Automation (ICRA)}, pages 16908--16914. IEEE, 2024.

\bibitem{dpo}
Rafael Rafailov, Archit Sharma, Eric Mitchell, Christopher~D Manning, Stefano Ermon, and Chelsea Finn.
\newblock Direct preference optimization: Your language model is secretly a reward model.
\newblock {\em Advances in Neural Information Processing Systems}, 36, 2024.

\bibitem{taskconditional}
Guolei Sun, Thomas Probst, Danda~Pani Paudel, Nikola Popovi{\'c}, Menelaos Kanakis, Jagruti Patel, Dengxin Dai, and Luc Van~Gool.
\newblock Task switching network for multi-task learning.
\newblock In {\em Proceedings of the IEEE/CVF international conference on computer vision}, pages 8291--8300, 2021.

\bibitem{multitaskdiffusionfinetuning}
Rabeeh~Karimi Mahabadi, Sebastian Ruder, Mostafa Dehghani, and James Henderson.
\newblock Parameter-efficient multi-task fine-tuning for transformers via shared hypernetworks.
\newblock {\em arXiv preprint arXiv:2106.04489}, 2021.

\bibitem{nuscenes}
Holger Caesar, Varun Bankiti, Alex~H Lang, Sourabh Vora, Venice~Erin Liong, Qiang Xu, Anush Krishnan, Yu~Pan, Giancarlo Baldan, and Oscar Beijbom.
\newblock nuscenes: A multimodal dataset for autonomous driving.
\newblock In {\em Proceedings of the IEEE/CVF conference on computer vision and pattern recognition}, pages 11621--11631, 2020.

\bibitem{rulebase1}
Elmar Brockfeld, Reinhart~D K{\"u}hne, Alexander Skabardonis, and Peter Wagner.
\newblock Toward benchmarking of microscopic traffic flow models.
\newblock {\em Transportation research record}, 1852(1):124--129, 2003.

\bibitem{rulebase2}
Pablo~Alvarez Lopez, Michael Behrisch, Laura Bieker-Walz, Jakob Erdmann, Yun-Pang Fl{\"o}tter{\"o}d, Robert Hilbrich, Leonhard L{\"u}cken, Johannes Rummel, Peter Wagner, and Evamarie Wie{\ss}ner.
\newblock Microscopic traffic simulation using sumo.
\newblock In {\em 2018 21st international conference on intelligent transportation systems (ITSC)}, pages 2575--2582. IEEE, 2018.

\bibitem{learningbase1}
Yuning Chai, Benjamin Sapp, Mayank Bansal, and Dragomir Anguelov.
\newblock Multipath: Multiple probabilistic anchor trajectory hypotheses for behavior prediction.
\newblock {\em arXiv preprint arXiv:1910.05449}, 2019.

\bibitem{learningbase2}
Yuxiao Chen, Boris Ivanovic, and Marco Pavone.
\newblock Scept: Scene-consistent, policy-based trajectory predictions for planning.
\newblock In {\em Proceedings of the IEEE/CVF Conference on Computer Vision and Pattern Recognition}, pages 17103--17112, 2022.

\bibitem{learningbase3}
Tim Salzmann, Boris Ivanovic, Punarjay Chakravarty, and Marco Pavone.
\newblock Trajectron++: Dynamically-feasible trajectory forecasting with heterogeneous data.
\newblock In {\em Computer Vision--ECCV 2020: 16th European Conference, Glasgow, UK, August 23--28, 2020, Proceedings, Part XVIII 16}, pages 683--700. Springer, 2020.

\bibitem{trace}
Davis Rempe, Zhengyi Luo, Xue Bin~Peng, Ye~Yuan, Kris Kitani, Karsten Kreis, Sanja Fidler, and Or~Litany.
\newblock Trace and pace: Controllable pedestrian animation via guided trajectory diffusion.
\newblock In {\em Proceedings of the IEEE/CVF Conference on Computer Vision and Pattern Recognition}, pages 13756--13766, 2023.

\bibitem{vae}
Diederik~P Kingma and Max Welling.
\newblock Auto-encoding variational bayes.
\newblock {\em arXiv preprint arXiv:1312.6114}, 2013.

\bibitem{diffusion}
Jonathan Ho, Ajay Jain, and Pieter Abbeel.
\newblock Denoising diffusion probabilistic models.
\newblock {\em Advances in neural information processing systems}, 33:6840--6851, 2020.

\bibitem{motiondiffuser}
Chiyu Jiang, Andre Cornman, Cheolho Park, Benjamin Sapp, Yin Zhou, Dragomir Anguelov, et~al.
\newblock Motiondiffuser: Controllable multi-agent motion prediction using diffusion.
\newblock In {\em Proceedings of the IEEE/CVF Conference on Computer Vision and Pattern Recognition}, pages 9644--9653, 2023.

\bibitem{traffcdiffusion1}
Zhiming Guo, Xing Gao, Jianlan Zhou, Xinyu Cai, and Botian Shi.
\newblock Scenedm: Scene-level multi-agent trajectory generation with consistent diffusion models.
\newblock {\em arXiv preprint arXiv:2311.15736}, 2023.

\bibitem{trafficdiffusion2}
Ethan Pronovost, Kai Wang, and Nick Roy.
\newblock Generating driving scenes with diffusion.
\newblock {\em arXiv preprint arXiv:2305.18452}, 2023.

\bibitem{trafficdiffusion3}
Ethan Pronovost, Meghana~Reddy Ganesina, Noureldin Hendy, Zeyu Wang, Andres Morales, Kai Wang, and Nick Roy.
\newblock Scenario diffusion: Controllable driving scenario generation with diffusion.
\newblock {\em Advances in Neural Information Processing Systems}, 36, 2024.

\bibitem{diffuser}
Michael Janner, Yilun Du, Joshua~B Tenenbaum, and Sergey Levine.
\newblock Planning with diffusion for flexible behavior synthesis.
\newblock {\em arXiv preprint arXiv:2205.09991}, 2022.

\bibitem{trafficdiffusion4}
Chejian Xu, Ding Zhao, Alberto Sangiovanni-Vincentelli, and Bo~Li.
\newblock Diffscene: Diffusion-based safety-critical scenario generation for autonomous vehicles.
\newblock In {\em The Second Workshop on New Frontiers in Adversarial Machine Learning}, 2023.

\bibitem{king}
Niklas Hanselmann, Katrin Renz, Kashyap Chitta, Apratim Bhattacharyya, and Andreas Geiger.
\newblock King: Generating safety-critical driving scenarios for robust imitation via kinematics gradients.
\newblock In {\em European Conference on Computer Vision}, pages 335--352. Springer, 2022.

\bibitem{navigate}
Matthew Niedoba, Jonathan Lavington, Yunpeng Liu, Vasileios Lioutas, Justice Sefas, Xiaoxuan Liang, Dylan Green, Setareh Dabiri, Berend Zwartsenberg, Adam Scibior, et~al.
\newblock A diffusion-model of joint interactive navigation.
\newblock {\em Advances in Neural Information Processing Systems}, 36, 2024.

\bibitem{behaviorgpt}
Zikang Zhou, Haibo Hu, Xinhong Chen, Jianping Wang, Nan Guan, Kui Wu, Yung-Hui Li, Yu-Kai Huang, and Chun~Jason Xue.
\newblock Behaviorgpt: Smart agent simulation for autonomous driving with next-patch prediction.
\newblock {\em arXiv preprint arXiv:2405.17372}, 2024.

\bibitem{xia2024language}
Junkai Xia, Chenxin Xu, Qingyao Xu, Chen Xie, Yanfeng Wang, and Siheng Chen.
\newblock Language-driven interactive traffic trajectory generation.
\newblock {\em arXiv preprint arXiv:2405.15388}, 2024.

\bibitem{fine-tuning1}
Nataniel Ruiz, Yuanzhen Li, Varun Jampani, Yael Pritch, Michael Rubinstein, and Kfir Aberman.
\newblock Dreambooth: Fine tuning text-to-image diffusion models for subject-driven generation.
\newblock In {\em Proceedings of the IEEE/CVF Conference on Computer Vision and Pattern Recognition}, pages 22500--22510, 2023.

\bibitem{fine-tuning2}
Rinon Gal, Yuval Alaluf, Yuval Atzmon, Or~Patashnik, Amit~H Bermano, Gal Chechik, and Daniel Cohen-Or.
\newblock An image is worth one word: Personalizing text-to-image generation using textual inversion.
\newblock {\em arXiv preprint arXiv:2208.01618}, 2022.

\bibitem{fine-tuning3}
Lvmin Zhang, Anyi Rao, and Maneesh Agrawala.
\newblock Adding conditional control to text-to-image diffusion models.
\newblock In {\em Proceedings of the IEEE/CVF International Conference on Computer Vision}, pages 3836--3847, 2023.

\bibitem{fine-tuning4}
Simo Ryu.
\newblock Low-rank adaptation for fast text-to-image diffusion fine-tuning, 2023.

\bibitem{fine-tuning5}
Ligong Han, Yinxiao Li, Han Zhang, Peyman Milanfar, Dimitris Metaxas, and Feng Yang.
\newblock Svdiff: Compact parameter space for diffusion fine-tuning.
\newblock {\em arXiv preprint arXiv:2303.11305}, 2023.

\bibitem{diffusionDPO}
Bram Wallace, Meihua Dang, Rafael Rafailov, Linqi Zhou, Aaron Lou, Senthil Purushwalkam, Stefano Ermon, Caiming Xiong, Shafiq Joty, and Nikhil Naik.
\newblock Diffusion model alignment using direct preference optimization.
\newblock {\em arXiv preprint arXiv:2311.12908}, 2023.

\bibitem{diffusionrlhf}
Zibin Dong, Yifu Yuan, Jianye Hao, Fei Ni, Yao Mu, Yan Zheng, Yujing Hu, Tangjie Lv, Changjie Fan, and Zhipeng Hu.
\newblock Aligndiff: Aligning diverse human preferences via behavior-customisable diffusion model.
\newblock {\em arXiv preprint arXiv:2310.02054}, 2023.

\bibitem{uehara2024feedback}
Masatoshi Uehara, Yulai Zhao, Kevin Black, Ehsan Hajiramezanali, Gabriele Scalia, Nathaniel~Lee Diamant, Alex~M Tseng, Sergey Levine, and Tommaso Biancalani.
\newblock Feedback efficient online fine-tuning of diffusion models.
\newblock {\em arXiv preprint arXiv:2402.16359}, 2024.

\bibitem{ctg}
Ziyuan Zhong, Davis Rempe, Danfei Xu, Yuxiao Chen, Sushant Veer, Tong Che, Baishakhi Ray, and Marco Pavone.
\newblock Guided conditional diffusion for controllable traffic simulation.
\newblock In {\em 2023 IEEE International Conference on Robotics and Automation (ICRA)}, pages 3560--3566. IEEE, 2023.

\bibitem{luca1988feedback}
A.~D. Luca, G.~Oriolo, and C.~Samson.
\newblock {\em Feedback Control of a Nonholonomic Car-Like Robot}, pages 171--249.
\newblock Springer, Berlin, Germany, 1988.

\bibitem{improved}
Alexander~Quinn Nichol and Prafulla Dhariwal.
\newblock Improved denoising diffusion probabilistic models.
\newblock In {\em International Conference on Machine Learning}, pages 8162--8171. PMLR, 2021.

\bibitem{video-diffusion-model}
Jonathan Ho, Tim Salimans, Alexey Gritsenko, William Chan, Mohammad Norouzi, and David~J. Fleet.
\newblock Video diffusion models.
\newblock {\em arXiv preprint arXiv:2204.03458}, 2022.

\bibitem{ditrefer}
Priya Goyal, Piotr Doll{\'a}r, Ross Girshick, Pieter Noordhuis, Lukasz Wesolowski, Aapo Kyrola, Andrew Tulloch, Yangqing Jia, and Kaiming He.
\newblock Accurate, large minibatch sgd: Training imagenet in 1 hour.
\newblock {\em arXiv preprint arXiv:1706.02677}, 2017.

\bibitem{dit}
William Peebles and Saining Xie.
\newblock Scalable diffusion models with transformers.
\newblock In {\em Proceedings of the IEEE/CVF International Conference on Computer Vision}, pages 4195--4205, 2023.

\bibitem{d3po}
Kai Yang, Jian Tao, Jiafei Lyu, Chunjiang Ge, Jiaxin Chen, Qimai Li, Weihan Shen, Xiaolong Zhu, and Xiu Li.
\newblock Using human feedback to fine-tune diffusion models without any reward model.
\newblock {\em arXiv preprint arXiv:2311.13231}, 2023.

\bibitem{du2024exploration}
Yihan Du, Anna Winnicki, Gal Dalal, Shie Mannor, and R~Srikant.
\newblock Exploration-driven policy optimization in rlhf: Theoretical insights on efficient data utilization.
\newblock {\em arXiv preprint arXiv:2402.10342}, 2024.

\bibitem{metricsrefer}
Danfei Xu, Yuxiao Chen, Boris Ivanovic, and Marco Pavone.
\newblock Bits: Bi-level imitation for traffic simulation.
\newblock In {\em 2023 IEEE International Conference on Robotics and Automation (ICRA)}, pages 2929--2936. IEEE, 2023.

\bibitem{strive}
Davis Rempe, Jonah Philion, Leonidas~J Guibas, Sanja Fidler, and Or~Litany.
\newblock Generating useful accident-prone driving scenarios via a learned traffic prior.
\newblock In {\em Proceedings of the IEEE/CVF Conference on Computer Vision and Pattern Recognition}, pages 17305--17315, 2022.

\end{thebibliography}

\end{document}